# Artificial Intelligence Enabled Material Behavior Prediction

Timothy Hanlon, Johan Reimann, Monica A. Soare, Anjali Singhal, James Grande, Marc Edgar, Kareem S. Aggour, and Joseph Vinciquerra

**Abstract**

Artificial Intelligence and Machine Learning algorithms have considerable potential to influence the prediction of material properties. Additive materials have a unique property prediction challenge in the form of surface roughness effects on fatigue behavior of structural components. Traditional approaches using finite element methods to calculate stress risers associated with additively built surfaces have been challenging due to the computational resources required, often taking over a day to calculate a single sample prediction. To address this performance challenge, Deep Learning has been employed to enable low cycle fatigue life prediction in additive materials in a matter of seconds.

**Introduction**

Artificial Intelligence and Machine Learning (AI/ML) is having a profound impact on the ways in which many organizations and businesses design, manufacture, monitor, inspect and repair industrial assets large and small. At GE, AI/ML is improving our ability to design, deliver and maintain world-class assets in the aerospace, energy, healthcare, and oil and gas industries, to name a few. For example, leveraging AI/ML to predict the mechanical behavior of materials represents a new frontier in the development of structural alloys. Considering the pace of development of additive materials, and the often rough surface conditions inherent to the process, ways to augment the prediction of low cycle fatigue (LCF) lives are explored.

The effect of surface roughness on the LCF behavior of metals and alloys has been studied for decades [1]. Surface preparation and machining techniques that reduce surface roughness, and those that leave the surface in a state of residual compression, have been successfully employed to extend fatigue lives of components that are subjected to cyclic stresses. While the same methodologies can be applied generally to the external surfaces of additively manufactured components, the internal 'as-built' surfaces of complex geometry parts can prove more challenging to modify. It is therefore critical to understand the surface characteristics of additively manufactured materials, and their subsequent effect on properties.

Traditionally, where LCF performance is a design requirement, surface roughness has been empirically accounted for via a correlation to crack initiation life, and its effect on endurance limit [2]. More recently, full finite element calculations of surface topography have been used to correlate the stress risers associated with microscopic surface asperities to fatigue life [3,4]. The computing resources required to make such calculations are considerable, however. Given the complexity of additively manufactured surfaces, these simulations become very extensive, both in terms of time and number of equations to be solved, to the point of being prohibitive. Machine Learning techniques such as Deep Learning are capable of significantly reducing the computational expense associated with these calculations. Deep Learning is a universal approximator [5] that is capable of building models over large amounts of data in very high dimensional space. We leverage advances in the Deep Learning space to build a surrogate model that can produce results similar to the finite element calculations, but with a significant reduction in processing time - from over a day to a few seconds per sample.

**Experimental Procedure**

In the application presented, additively manufactured coupons were non-destructively interrogated via nano-computed tomography (nano-CT), with a pixel resolution of 3 microns. The resulting nano-CT images were then analyzed, meshed, and finite element calculations were run to assess the stress concentrations associated with surface asperities.

Figure 1a illustrates a reconstructed 3D image of an additively manufactured hollow cylindrical LCF specimen, with an as-built inner diameter surface. A radial-axial slice within that volume is shown in Figure 1b. The asperities associated with the natural roughness of the as-built surface represent local stress concentrations, which are expected to accelerate the crack initiation and early crack growth processes, thereby reducing the total LCF life of the part.

To evaluate the effect of surface roughness on life, nano-CT images taken from the gage section of LCF specimens were reconstructed and meshed, and stress concentration factors ($K_t$) were calculated locally using a finite element structural elastic analysis. $K_t$ values obtained in a full 3D simulation were compared with those extracted from a 2D analysis. In the latter, 720 2D nano-CT radial-axial sections, equally distributed around the circumference, were reconstructed and meshed over the same volume examined in the 3D simulation.

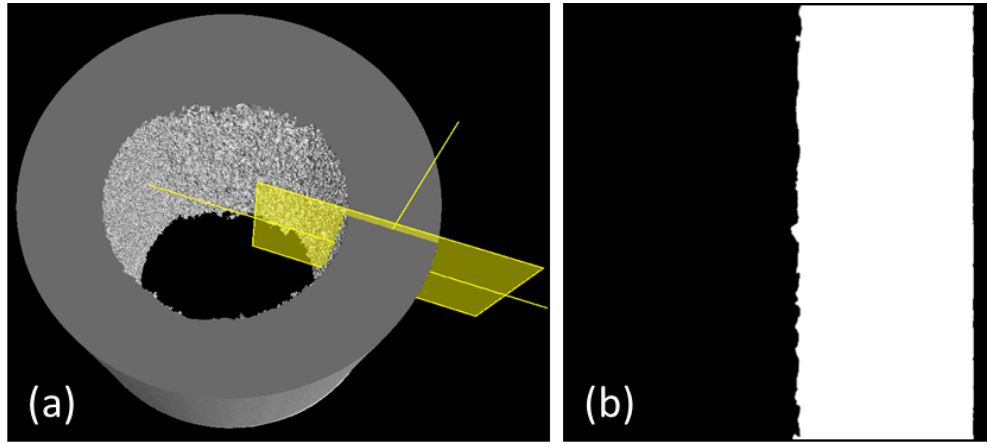

Figure 1. Geometrical representation of a cylindrical LCF bar, reconstructed from nano-CT measurements. a) 3D representation, b) 2D radial-axial section.

Figure 2a shows the first principal stress distribution on a 3D cylinder with a length of ~1mm. The cylinder was meshed with over 3.6 million tetrahedral elements (9 microns on average in size), optimized to capture the geometry of the surface features. Some of the original nano-CT resolution was lost in this process, but the optimized unstructured mesh allowed for a very good representation of the surface asperities, eliminating the typical surface steps from a structural mesh and minimizing the number of elements. Figure 2b shows the first principal stress in one of the 720 sections, meshed with structured 2D axisymmetric elements (square shape). The initial size of each element is 3 microns, maintaining exactly the resolution used in the nano-CT scans. The image in Figure 2b contains over 120,000 elements covering a section of approximately 1mm in the axial direction. The exceedance probability function for the volumes having a stress concentration factor greater than 2.5 was found to have the same variation in the 3D case as in the 2D case for axial loading conditions (Figure 2c). This implies that the 2D analysis is sufficient to calculate the local $K_t$ distributions of interest in this case. In terms of the finite element analysis simulation/post processing time, there is no significant difference between a full 3D simulation and 720 2D simulations for the same volume. In this case, both took ~24 hours on a 4-core machine. However, the pre-processing time associated with meshing the nano-CT images is longer for the 3D volume (a few hours), relative to the 720 2D sections (less than an hour). Irrespective of the analysis type (2D vs. 3D) the total processing time for each specimen gage section is longer than 24 hours.

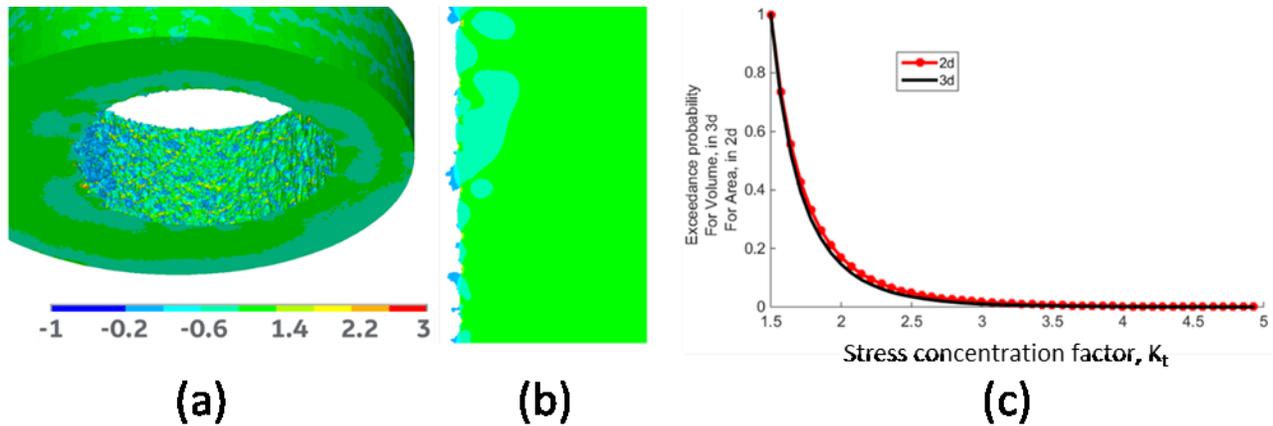

Figure 2. Stress concentration factor distribution, as calculated using a) 3D analysis and b) 2D axisymmetric analysis. c) Exceedance probability function for the volume affected by a $K_t$ greater than 2.5.

Because of the cost and time required to run such simulations, a Deep Learning algorithm was developed to predict the same information in a fraction of the time (i.e. seconds). To do so, a convolutional network was first trained on a series of finite element calculations that mimic the general range of surface asperity geometries expected in the part. Figure 3 shows a comparison of the first principal stress, as calculated using the finite element method versus predicted using the Deep Learning algorithm. The error is less than 5%. Once the Deep Learning algorithm is trained over the range of expected surface roughness, it can be applied over large area scans of the material to produce stress concentration maps on the part, as in Figure 4. Such large area scans would ordinarily be too computationally intensive to run with conventional finite element analyses on a routine basis, making the AI/ML approach attractive for this type of estimate.

The Deep Learning model was designed to model the expected physical properties of the stress concentrations, that is, if stress propagation was modified by macro features such as interactions between surface features that are not near one another, the Deep Learning model is able to capture this context and modify the stress concentration estimate accordingly.

The network itself consists of 7 convolutional layers of compression and 7 deconvolutional layers of decompression, each with rectified linear units as their activation function apart from the output layer, whose activation function is linear (Figure 3d). For the sample sizes analyzed in this work, this level of compression allowed the system to capture the interplay between the edges of the image.

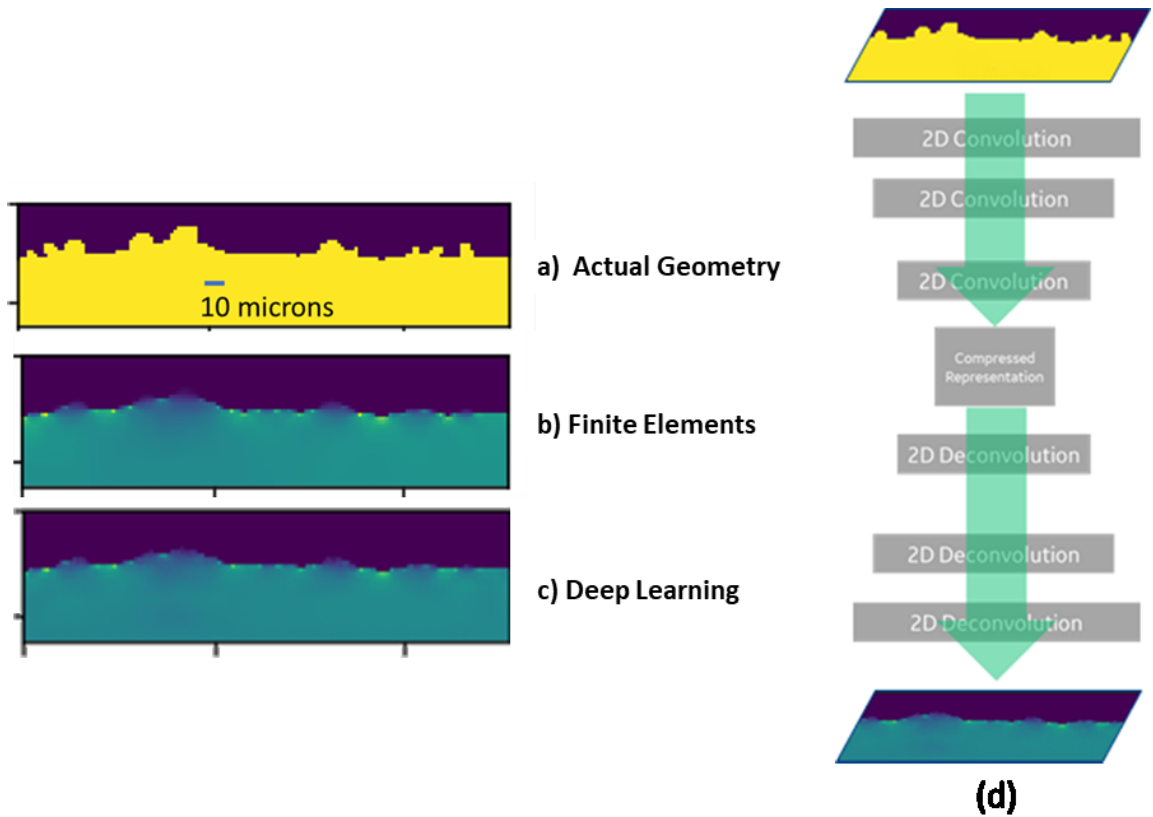

Figure 3. (a) Actual surface topography as measured via nano-CT. (b) Finite element calculation of stress concentrations associated with surface asperities. (c) Deep Learning prediction of stress concentrations associated with surface asperities. (d) Illustration of the network architecture used to generate the stress concentration estimate. The compressed representation captures the local and global effects that impact the estimate.

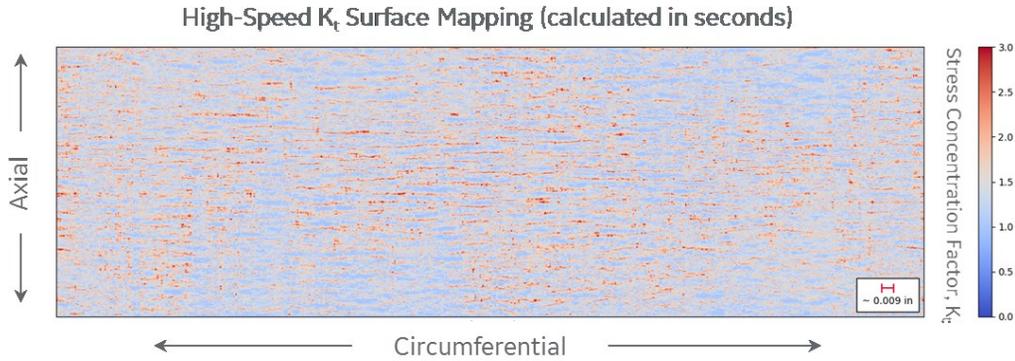

Figure 4. Large area stress concentrations calculated using the Deep Learning algorithm.

**Results and Discussion**

With the fast-acting predictive capability afforded by the ML model, it becomes possible to explicitly add surface roughness effects to the lifing strategy of choice. Because crack initiation is often related to the near-surface stressed volume effects, an algorithm was developed to assess the volume adjacent to surface asperity features that exceeded a stress concentration factor ($K_t$) of 2.5. This threshold was chosen to signify a significant elevation in local stress near the surface, though any user-specified value could be chosen. An image-processed nano-CT slice of an additively built surface is represented in Figure 5a. The results of the ML model used to predict $K_t$ are shown in Figure 5b. Figure 5c represents the output of the algorithm designed to identify clusters of neighboring pixels that exceed the $K_t$ threshold of 2.5. This represents an attempt to understand how the stressed volume at the surface changes with varying additive manufacturing build parameters and subsequent native surface roughness. Samples with larger cluster sizes and higher cluster number densities are assumed to be at higher risk of lower LCF lives, given the often-stochastic nature of crack initiation. Grains favorably oriented for deformation have a higher likelihood of residing in a high $K_t$ region and initiating a crack the more stressed volume that exists in the material. Figure 5d indicates the distribution of neighboring pixel clusters predicted to have $K_t$ greater than 2.5. Here, cluster size represents the number of neighboring pixels that meet this Kt threshold. Figure 5e represents an early attempt to utilize this information, along with test conditions, to predict the LCF lives across several additively manufactured coupons. While the predictions fall within the ~2x scatter band typically observed experimentally in LCF testing, there are several improvements planned to incorporate additional aspects of the underlying microstructure in this lifing model. Specifically, ML tools will be designed to extract quantitative grain size information, phase distributions, retained strain and grain orientation/texture information, among others. That collection of data will then also feed the life prediction tool to improve its fidelity.

Additionally, since the accuracy of these predictions requires application of the method over many tens of fatigue bars, Artificial Intelligence methods can reduce the analysis time from months to days, relative to brute force finite element calculations over the same volume of material. This brings substantial cost savings to the design cycle of a part. The methodology can also be considered as an in-line quality control mechanism during production.

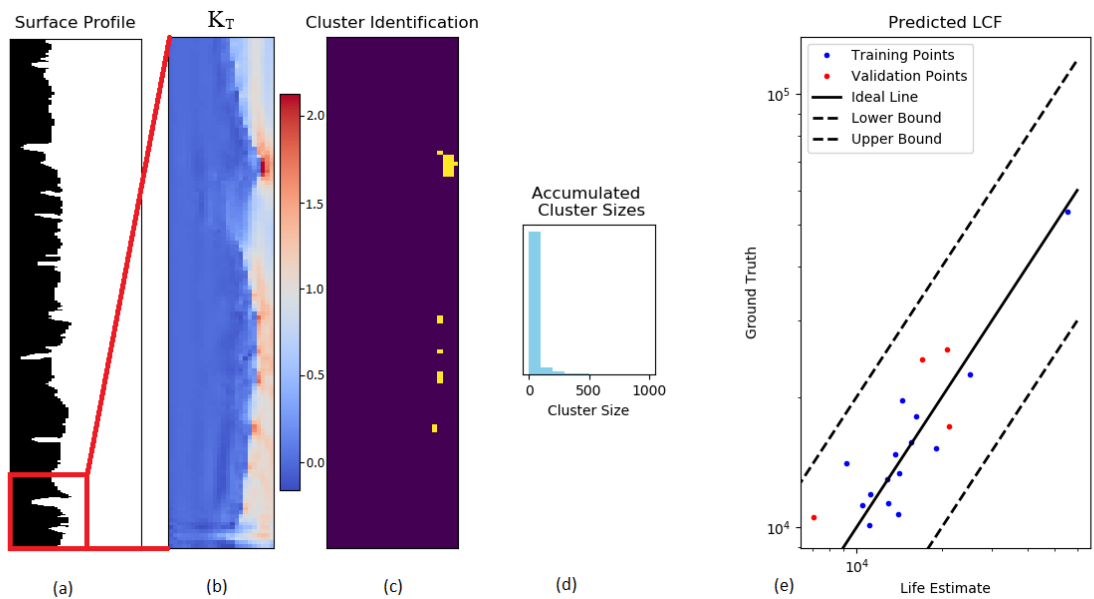

Figure 5. Process of converting images of surface roughness to stress concentration profiles. Clusters of pixels with elevated stress levels can then be quantified to simulate stressed volume, which can be applied to augment lifing strategies.

**Conclusions and Future Work**

Machine Learning algorithms can be leveraged to expedite the calculations required to improve the prediction of material properties. This represents an incremental step toward broader adoption of these practices in the lifing methodology community. Here, Machine Learning was used to accelerate the prediction of local stress concentration factors resulting from the inherent surface roughness of additively manufactured components. An extension currently being explored is the use of Machine Learning to extract and quantify critical features of the underlying microstructure that also affect material behavior. The output will inform life models already under development, improving their fidelity even further.